\documentclass{article} 
\usepackage{nips12submit_e,times}

\usepackage{graphicx}
\usepackage{subfigure}

\usepackage{algorithm}
\usepackage{algorithmic}

\usepackage{latexsym}
\usepackage{amsmath}
\usepackage{amsfonts}
\usepackage{multirow}
\usepackage{url}
\usepackage{verbatim}

\title{Kernelized Locality-Sensitive Hashing for Semi-Supervised Agglomerative Clustering}

\author{
Boyi Xie \\
Department of Computer Science\\
Columbia University\\
New York, NY 10027 \\
\texttt{xie@cs.columbia.edu} \\
\And
Shuheng Zheng \\
Department of Industrial Engineering \& Operations Research \\
Columbia University\\
New York, NY 10027 \\
\texttt{sz2228@columbia.edu} \\
}

\nipsfinalcopy 

\begin{document}

\maketitle

\begin{abstract}
Large scale agglomerative clustering is hindered by computational burdens. We propose a novel scheme where exact inter-instance distance calculation is replaced by the Hamming distance between Kernelized Locality-Sensitive Hashing (KLSH) hashed values. This results in a method that drastically decreases computation time. Additionally, we take advantage of certain labeled data points via distance metric learning to achieve a competitive precision and recall comparing to K-Means but in much less computation time.
\end{abstract}

\section{Introduction}
\label{introduction}

Our proposed research topic is to do clustering on a scalable dataset from a semi-supervised approach based on hashing methods. 
In particular, our goal is to explore the underlying data distribution by clustering the data points and differentiating the classes. When a small set of labeled data that come from only a subset of the classes is given, we want to find out the whole data distribution for a complete set of classes. For example, we are given a set of labels of two classes, can we separate these two classes well and at the same time discover the existence of a third class. It requires using the information from the labeled data to find a transformation metric that can split the two classes well; and after this data transformation, we can discover that there is a third class exists. Suppose there is a handwritten digit recognition task and the dataset contains digits `2', `7' and `4'. If a general agglomerative clustering is run, it might end up with 2 clusters that `2' and `7' in one cluster and `4' in the other, due to the similarity of their shapes. However, when a small labeled set of classes `2' and `7' is given, we can learned a degree of granularity for similarity comparison. By using a data transformation that maximally can split `2' and `7' into two clusters, we are able to identify the existence of another cluster, digit `4'. Because agglomerative clustering suffers from its computation inefficiency, a major contribution of this paper is to introduce a machine learned hashing method - kernelized locality-sensitive hashing (KLSH) - into agglomerative clustering. This results in an efficient computation in clustering for large-scale dataset.

Our paper is structured as follows. We provide background study and related work in section 2. Section 3 presents our algorithms for distance metric learning and KLSH clustering. Section 4 describes the experiments with a discussion of the results, followed by conclusions in section 5.

\section{Related Work}

There has been much previous work on cluster seeding to address the limitation that iterative clustering techniques (e.g. K-Means and Expectation Maximization (EM)) are sensitive to the choice of initial starting points (seeds). The problem addressed is how to select seed points in the absence of prior knowledge. Kaufman and Rousseeuw \cite{Kaufman90} propose an elaborate mechanism: the first seed is the instance that is most central in the data; the rest of the representatives are selected by choosing instances that promise to be closer to more of the remaining instances. Pena et al. \cite{Pena99} empirically compare the four initialization methods for the K-Means algorithm and illustrate that the random and Kaufman initializations outperform the other two, since they make K-Means less dependent on the initial choice of seeds. In K-Means++ \cite{Arthur07}, the random starting points are chosen with specific probabilities: that is, a point $p$ is chosen as a seed with probability proportional to $p$'s contribution to the overall potential (defined by the sum of  squared distances between each point and the closest center). By augmenting K-Means using this simple, randomized seeding technique, K-Means++ is $\theta$(log K) competitive with the optimal clustering. Bradley and Fayyad \cite{Bradley98} propose refining the initial seeds by taking into account the modes of the underlying distribution. This refined initial seed enables the iterative algorithm to converge to a better local minimum.
Semi-supervised learning is also seen as unsupervised learning guided by constraints. Noticed that clustering is heavily dependent on distance metrics and a particular algorithm is an executor to follow the rules, \cite{Xing02} pointed out the desire to use a systematic way to learn distance metric for clustering from labeled data. It is based on posing metric learning as a convex optimization problem. 

When the data size is growing exponentially, hashing is a technique especially good at solving large scale problems. \cite{Andoni08} described Locality-Sensitive Hashing (LSH) method, which is an efficient algorithm for the approximate and exact nearest neighbor problem. Their goal is to preprocess a dataset of objects (e.g. images) so that later, given a new query object, one can quickly return the dataset object that is most similar to the query. The technique is of significant interest in a wide variety of areas of unsupervised learning. Hierarchical clustering tries to solve a similar problem, from another perspective. By iteratively finding nearest neighbors, it groups data into clusters. Kernelized LSH is later proposed by \cite{Kulis09} for fast image search. It generalizes LSH to accommodate arbitrary kernel functions, making it possible to preserve the algorithm's sub-linear time similarity search guarantees for a wide class of useful similarity functions.

\section{Methods}

In this section, we describe our methods to solve a large scale semi-supervised learning problem by first introducing the distance learning metrics, and then our fast agglomerative clustering method based on kernelized locality-sensitive hashing (KLSH). 

\subsection{Distance Metric Learning}

Under the circumstances that the data given to us has a few labeled points, and we know which points for sure belongs to the same or different classes. We have a similarity and a dissimilarity matrix $S$ and $D$ respectively. For entry $s_{i,j}$ in similarity matrix $S$, $s_{i,j}=1$ if data $x_{i}$ and $x_{j}$ are in the same class and $0$ otherwise. Similarly for dissimilarity matrix $D$. Based on \cite{Xing02}, we try to learn a distance metric $|| x - y || _{A} = \sqrt {(x - y) ^{T} A (x - y)}$, where $x$, $y$ are two data points, $A$ is a positive semi-definite matrix of distance parameters among data points. The idea is to minimize the distance between similar points while keeping dissimilar points apart.

\begin{align*}
min & \sum_{(x_{i},x_{j}) \in S} || x_{i} - x_{j} || _{A}^{2} \\
& \sum_{(x_{i},x_{j}) \in D} || x_{i} - x_{j} || _{A} \geq 1 \\
& A \succeq 0
\end{align*}

It can be solved efficiently using constrained Newton's descent on the objective function

\[
\sum_{(x_{i},x_{j}) \in S} || x_{i} - x_{j} || _{A}^{2} - log(\sum_{(x_{i},x_{j}) \in D} || x_{i} - x_{j} || _{A}).
\]

This semi-supervised part is to learn a distance metric for data transformation before the main agglomerative clustering.

\subsection{Clustering with KLSH}

Curse of dimensionality is a well-known problem for learning on large scale datasets. It is related to the fact that the cost of computation grows exponentially with the increase of data dimensions, or the number of data instances. This is a problem directly affects clustering approaches that based on density estimation in input space.

For instance, in K-Means or general agglomerative clustering, the cost in iteratively estimating new centroid locations and re-arranging data instances to clusters exerts a significant burden on the performance. This happens especially to dataset in high dimension, where frequently computing inter-instance distances are highly expensive.

Our proposed semi-supervised clustering algorithm using kernelized locality-sensitive hashing (KLSH) in Algorithm~\ref{alg:ssch} aims to solve the large scale agglomerative clustering problem. It first learn a distance metric $A$ from a small set of labeled data (step 1 in Algorithm \ref{alg:ssch}). The second step is to build KLSH table that map the data in to hashed bits. In the rest of the procedure, an agglomerative clustering is performed.
Instead of explicitly computing inter-instance distances, the clustering is done based on the KLSH-hashed data points by measuring their Hamming distance. Such kernelized locality-sensitive hashing method has a high probability of preserving neighborhoods so it's a reasonable substitute for the exact inter-instance distances.

\begin{algorithm}[tb]
   \caption{Semi-Supervised Clustering with Hashing}
   \label{alg:ssch}
\begin{algorithmic}
   \STATE {\bfseries Input:} DataLabeled(limited), Data $x_1,...,x_N$.
   \STATE {\bfseries Step1:} Learn distance metric \textit{A} from labeled (limited) data
   \STATE {\bfseries Step2:} Build \textit{A}-Distance KLSH table (Algorithm~\ref{alg:klsh}).
   \STATE {\bfseries Step3:}
   \STATE {\bfseries Initialization:}
   \STATE $\bullet$ Let cluster distribution $R_0=\{\{x_i\},i=1,...,N\}$, i.e. each data point is an individual cluster $C_i$.
   \STATE $\bullet$ Let proximity matrix $P_0=P(X)$ of hash keys.
   \STATE $\bullet$ Set $t=1$.
   \REPEAT
   \STATE $\bullet$ $t=t+1$
   \STATE $\bullet$ Find $C_i$, $C_j$ such that $d(C_i,C_j)=min_{r,s=1,...,N,r \neq s}d(C_r,C_s)$.
   \STATE $\bullet$ Merge $C_i$, $C_j$ into a single cluster $C_q$ and form $R_t=(R_{t-1}-{C_i,C_j}) \cup {C_q}$.
   \STATE $\bullet$ Define the proximity matrix $P_t$ from $P_{t-1}$ by (a) deleting the two rows and columns that corresponding to the merged clusters and (b) adding a row and a column of the new cluster.
   \UNTIL{(The remaining number of clusters is equal to a specified $k$; or the inconsistency coefficient exceeds a threshold.)}
   \STATE {\bfseries Step4:} Retrieve actual data instances from KLSH hash table for the corresponding clusters
\end{algorithmic}
\end{algorithm}

\begin{algorithm}[tb]
   \caption{Build \textit{A}-Distance KLSH Table}
   \label{alg:klsh}
\begin{algorithmic}
   \STATE {\bfseries Input:} Data $x_i,...,x_N$, distance parameter $A$.
   \STATE {\bfseries Step1:} Randomly select p points from data, denoted as $x_1,...,x_i,...,x_p$.
   \STATE Build kernel $K(i,j)=exp(-(d(x_i,x_j)^2/\sigma^2)$, for $i,j=1,...,p$, where $d(x,y)=\sqrt{(x-y)'A(x-y)}$.
   \STATE {\bfseries Step2:} Apply SVD to $K$, suppose $K=U \Sigma U^T$. $K^{-1/2}=U\Sigma^{-1/2}U^T$.
   \STATE {\bfseries Step3:} Form a $p$-dim vector $e_s$, where $t$ dimensions are $1$ while others are $0$. These $t$ dimensions are chosen randomly.
   \STATE {\bfseries Step4:} $w=K^{-1/2}e_s$. For any $x$, the bit is created as $h(x)=sign(\Sigma_{i=1}^p w_i k(x,x_i))$.
\end{algorithmic}
\end{algorithm}

\section{Experiments} 

Our experiment is based on the MNIST dataset of handwritten digits. We evaluate our KLSH agglomerative clustering algorithm via a comparison to K-Means. 

\subsection{Datasets}

We obtained handwritten digits from the MNIST data repository. There are 10 classes of rasterized images (corresponding to digits from `0' to `9'). We used up to 50,000 data points for experiments.

\subsection{Experiment Setup}

We ran experiments using both K-Means and KLSH agglomerative clustering with and without distance metric learning. Hash string length, the number of classes, and the number of data points are varied one at a time. We report precision, recall and the computation time. All the experiments were done on a machine with 8-core Intel processors of 2.8 GHz and 8 GB of RAM.

\subsection{Results and Analysis}

Tables \ref{result}-\ref{result_bits} summarize our results, and the followings are several trends to notice.

First in Table \ref{result}, we observed that KLSH agglomerative clustering can achieve the same level of precision for a fraction of the computational cost. The downside is that recall is caused by the factor of 2. The decrease in recall is caused by the fact that KLSH cannot recover all of the points in the nearest neighborhood. The addition of distance metric learning has noticeable benefits on performance for KLSH Agglomerative Clustering.

In Table \ref{result_classes}, we analyzed the effect of an increase in the number of classes (while fixing the number of data points) on precision, recall, and computation time. Precision remains constant while recall decreases. The computational costs remains relatively independent of the number of clusters.

In Table \ref{result_bits}, we analyzed the effect of hash string length on clustering validity. Increasing the length of hash string increases both the precision and recall.

It is also able to adjust the tradeoff between efficiency and effectiveness. Notice that even if we use $m=32$ bit binary hash code, there are still $2^{32}$ possible outcomes. If the hashing split data well, the number of entries of the table will still be very large. It increases the accuracy of clustering results but meanwhile leads to a higher computation cost during agglomerative clustering.

According to the results, clustering with KLSH has superior performance when the dataset is large and the number of real clusters is small. Comparing to K-Means, it has large promising improvement on speed. When true cluster number is not large, it achieves high performance on both speed and accuracy. Especially in a lower level of the linkage tree, clustering with bias (distance metric learning) can immediately correctly cluster similar data instances.

\begin{table*}[tbp]
\caption{Experiment results that compare four methods: (1) K-Means, (2) K-Means with Distance-metric Learning, (3) Agglomerative clustering using KLSH and (4) Agglomerative clustering using KLSH with Distance-metric Learning. Precision, recall and computation time is reported. They all run on data underlying 10 classes and the hash code is 32-bit.}
\label{result}
\vskip 0.15in
\begin{center}
\begin{small}
\begin{sc}
\begin{tabular}{ | p{0.29 in} || p{0.17 in} | p{0.17 in} | p{0.37 in} || p{0.17 in} | p{0.17 in} | p{0.37 in} || p{0.17 in} | p{0.17 in} | p{0.37 in} || p{0.17 in} | p{0.17 in} | p{0.28 in} || } \hline
	 \multirow{2}{*}{\# Inst.}  & \multicolumn{3}{|c||}{K-Means} & \multicolumn{3}{|c||}{K-Means w/ DL} & \multicolumn{3}{|c||}{Aggl. KLSH} & \multicolumn{3}{|c||}{Aggl. KLSH w/ DL}     \\ \cline{2-13} 
     &    Pre &   Rec &   Time  &    Pre &   Rec &   Time  &    Pre &   Rec &   Time  &    Pre &   Rec &   Time   \\ \hline
\hline
5000   &  .590 & .564 & 13.246 & .537 & .496  & 15.647  & .573 & .305 & 2.155 & .631 & .272 & 2.466  \\ \hline
10000 &  .568 & .540 & 46.398 & .580 & .556  & 33.736  & .520 & .336 & 7.255 & .613 & .250 & 5.246  \\ \hline
15000 & .574  & .530 & 69.252 & .556 & .539 & 186.469 & .584 & .180 & 13.843 & .610 & .156 & 8.077 \\ \hline
20000 & .589 & .563 & 79.499 & .455 & .448 & 112.178 & .609 & .355 & 3.052 & .617 & .292 & 18.070 \\ \hline
30000 & .523 & .503 & 164.853 & .552 & .541 & 139.773 & .624 & .235 & 58.646 & .548 & .306 & 23.136 \\ \hline
50000 & .560 & .531 & 339.599 & .565 & .530 & 333.313 & .579 & .230 & 126.280 & .590 & .252 & 122.558 \\ \hline
\end{tabular}
\end{sc}
\end{small}
\end{center}
\vskip -0.1in
\end{table*}

\begin{table*}[tbp]
\caption{Compare the performance of various number of underlying classes for agglomerative clustering using KLSH with distance metric learning. In this case, data size is 20,000 and the hash code is 32-bit. With an increase in the number of classes, the precision remains constant while recall decreases}
\label{result_classes}
\vskip 0.15in
\begin{center}
\begin{small}
\begin{sc}
\begin{tabular}{ | c || c | c | c |} \hline
	 \# Classes  & Pre & Rec & Time    \\ \hline
4  & .714 & .465 & 16.527 \\ \hline
5  & .629 & .355 & 19.226 \\ \hline
6  & .702 & .507 & 18.747 \\ \hline
7  & .611 & .317 & 21.675 \\ \hline
8  & .654 & .354 &21.540 \\ \hline
9  & .603 & .284 & 24.024 \\ \hline
10 & .617 & .292 & 18.070 \\ \hline
\end{tabular}
\end{sc}
\end{small}
\end{center}
\vskip -0.1in
\end{table*}

\begin{table*}[tbp]
\caption{Compare the performance of various number of hash code bits for agglomerative clustering using KLSH with distance metric learning. In this case, data size is 20,000 with underlying 10 classes. Increasing the length of hash string increases both the precision and recall.}
\label{result_bits}
\vskip 0.15in
\begin{center}
\begin{small}
\begin{sc}
\begin{tabular}{ | c || c | c | c |} \hline
	 \# Bits  & Pre & Rec & Time    \\ \hline
8  & .402 & .245 & 0.043 \\ \hline
16 & .599 & .111 & 1.000 \\ \hline
32 & .617 & .292 & 18.070 \\ \hline
64 & .635 & .380 & 92.452 \\ \hline
\end{tabular}
\end{sc}
\end{small}
\end{center}
\vskip -0.1in
\end{table*}

\section{Conclusions}

General hierarchical clustering methods cannot scale well on large dataset due to the exponentially growing number of calculations on inter-instance distances. 
Kernelized locality-sensitive hashing (KLSH) provides a high probability of preserving neighborhoods and it's a reasonable substitute for the exact inter-instance distances. 
Our proposed KLSH agglomerative clustering alleviates the problem by calculating a reduced-sized Hamming distance and achieves efficient clustering computation. 
The incorporation of 
distance metric learning marginally improves the precision and recall.

\bibliographystyle{unsrt}
\bibliography{klshclustering}

\end{document}